\documentclass[letterpaper]{article} 
\usepackage{aaai23}  
\usepackage{times}  
\usepackage{helvet}  
\usepackage{courier}  
\usepackage[hyphens]{url}  
\usepackage{graphicx} 
\usepackage{natbib}  
\usepackage{caption} 
\usepackage{algorithm}
\usepackage{algorithmic}

\usepackage{newfloat}
\usepackage{listings}

\usepackage{algorithm}
\usepackage{algorithmic}
\usepackage{amsmath}

\DeclareCaptionStyle{ruled}{labelfont=normalfont,labelsep=colon,strut=off} 
\floatstyle{ruled}
\newfloat{listing}{tb}{lst}{}
\floatname{listing}{Listing}
\title{Semi-Supervised Junction Tree Variational
Autoencoder for Molecular Graphs}
\author{
    Atia Hamidizadeh,
    Tony Shen,
    Martin Ester
}
\affiliations{
    School of Computing Science, Simon Fraser University\\

    8888 University Drive, Burnaby, British Columbia\\
    \{ahamidiz, tsa87, ester\}@sfu.ca
}

\usepackage{bibentry}

\begin{document}

\maketitle

\begin{abstract}
Molecular Representation Learning is essential to solving many drug discovery and computational chemistry problems. It is a challenging problem due to the complex structure of molecules and the vast chemical space. Graph representations of molecules are more expressive than traditional representations, such as molecular fingerprints. Therefore, they can improve the performance of machine learning models. We propose SeMole, a method that augments the Junction Tree Variational Autoencoders, a state-of-the-art generative model for molecular graphs, with semi-supervised learning. SeMole aims to improve the accuracy of molecular property prediction when having limited labeled data by exploiting unlabeled data. We enforce that the model generates molecular graphs conditioned on target properties by incorporating the property into the latent representation. We propose an additional pre-training phase to improve the training process for our semi-supervised generative model. We perform an experimental evaluation on the ZINC dataset using three different molecule properties and demonstrate the benefits of semi-supervision. 
\end{abstract}

\section{Introduction}

One of the challenges in the drug development pipeline is to discover small molecules with desired properties. Testing candidate molecules experimentally in the wet lab is time-consuming and expensive. The main challenge for computational methods is the vast chemical space and the challenging nature of navigating through this space where molecule representation learning attracts attention. Molecule representation learning has been utilized by either end-to-end training or pretrained strategies to solve drug discovery problems such as generating molecules \cite{jin2018junction, dai2018syntax,kusner2017grammar}, and molecule property prediction \cite{gilmer2017neural, yang2019analyzing, hu2019strategies}. Recently, advancements in learning molecule representations have been made to propose novel molecules with targeted desired properties \cite{gomez2018automatic, kang2018conditional, olivecrona2017molecular, popova2018deep}

\begin{figure*}[htbp]
\centerline{\includegraphics[scale=0.6]{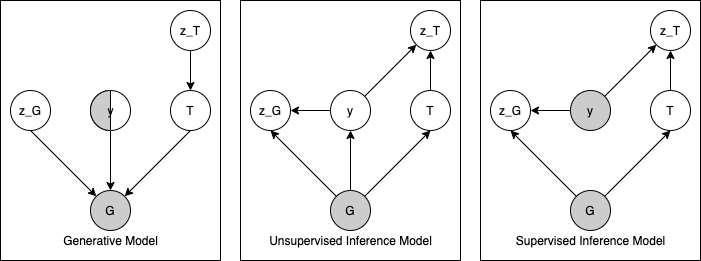}}
\caption{Probabilistic semi-supervised model on molecules represented by graphs and trees. \textbf{Left}: Generative Model. \textbf{Middle}: Unsupervised Inference Model. \textbf{Right}: Supervised Inference Model.}
\label{fig1}
\end{figure*}

Several molecular representation learning models have been designed by representing molecules as graphs, where nodes are atoms and bonds are edges. Utilizing Graph Neural Networks(GNNs) \cite{wu2020comprehensive} have resulted in promising performances for unconditionally generating molecules \cite{jin2018junction,ma2018constrained,you2018graph,shi2020graphaf,you2018graphrnn}. Junction Tree Variational Autoencoders(JTVAE) \cite{jin2018junction} is one of a state-of-the-art graph-based methods for generating molecules. This method converts the graph structure to the associated tree structure of molecules by breaking them into components predefined in the vocabulary of chemical bonds. This ensures the model generates molecules by assembling chemically valid blocks that result in generating 100\% valid molecules.

Previous generative methods do not generate molecules conditioned on target properties, but optimize the latent space based on a target property. \cite{jin2018junction},\cite{kusner2017grammar}, and \cite{dai2018syntax} learn the latent representation in an unsupervised manner, minimizing the reconstruction error, and optimize molecules with respect to the desired property afterward, following the optimization approach proposed by \cite{gomez2018automatic}. Despite these methods' promising performance, the optimization approaches do not allow setting the property to a specific value and depend on the definition of the objective function for the optimization task. Therefore They are also not scalable for generating molecules with multiple desired properties. 

Moreover, these supervised training methods assume that there is enough labeled data to train a classifier or regressor. These solutions largely depend on the availability of labeled datasets, which is not the case in many real-world datasets. Recently, researchers have proposed methods for semi-supervised graph classification \cite{sun2019infograph, hao2020asgn} to solve the data scarcity problem. Therefore semi-supervised learning could be beneficial in enhancing the power of molecule generative models. 

Kang et al.\cite{kang2018conditional} proposed a semi-supervised variational autoencoder(SSVAE) that conditionally generates molecules without any post hoc optimization. However, SSVAE represents molecules as SMILES strings \cite{weininger1988smiles}, which often leads the model into generating invalid molecules. The SMILES representation of molecules is also not sensitive to molecule similarities, making it hard to learn a smooth embedding of molecules \cite{jin2018junction}. Moreover, semi-supervised learning with generative models \cite{kingma2014semi} is challenging to train end-to-end \cite{maaloe2016auxiliary}.

In this paper, we extend a state-of-the-art generative model, JTVAE, for molecular graphs with semi-supervised learning to learn molecular properties directly as part of the latent representations via partial supervision. We propose an additional pre-training phase to improve the training process for the semi-supervised generative model. While JTVAE achieved state-of-the-art performance on generating molecules unconditionally, generating molecules conditioned on target properties with limited labeled data remains largely untapped. In conclusion, this paper makes the following contributions:

\begin{itemize}
\item We combine a semi-supervised model with a state-of-the-art molecular graph generative model, JTVAE, for molecule property prediction and generation of valid molecules with desired properties. 
\item We improve the  training process of Semi-supervised Variational Autoencoder by adding a pre-training phase. 
\item We performed an experimental evaluation on ZINC dataset for three different molecule properties on molecule property prediction and generation with respect to target properties.
\end{itemize}

\section{Related Work}

\subsection{Semi-supervised Learning for Molecular Graphs}

Semi-supervised learning is particularly beneficial for chemistry applications due to the vast unlabeled chemical space and frequently available partially labeled data in the pharmaceutical and precision agriculture industry.
ASGN\cite{hao2020asgn} addressed the data scarcity problem using a semi-supervised method that predicts the most informative samples obtained by an active learning approach. InfoGraph\cite{sun2019infograph} is a semi-supervised graph-level representation learning method that has been evaluated on molecular property prediction. This method employs a student-teacher framework where the teacher model is trained on unlabeled data, and the student model is trained on the labeled data using a supervised objective. InfoGraph maximizes the mutual information between the representations learned by these two models so that the student model learns from the teacher model. 

Semi-supervised learning using the generative models optimizes the prediction jointly with a Variational Autoencoder over the input data \cite{kingma2014semi, maaloe2016auxiliary, siddharth2017learning}. The latent representation is divided into a structured and unstructured part, where the structured part is enforced to represent labels of data points. Only part of the labels is provided during the training; therefore, the model learns the latent representation in a semi-supervised setting. This approach has achieved state-of-the-art performance on image classification \cite{kingma2014semi, maaloe2016auxiliary} and speech synthesis \cite{habib2019semi} with partially labeled data\cite{kingma2019introduction}. \cite{kang2018conditional} proposed a method for conditionally generating molecules inspired by semi-supervised learning with generative models \cite{kingma2014semi}.

\section{Methodology}

In this section, we first introduce our problem definition. Then we propose our method, SeMole, for semi-supervised Junction Tree Variational Autoencoder. Afterward,  we present a modified version by adding a pretraining phase to training, which we call SeMole$_{Pretrained}$.

\subsection{Problem Definition}

Given a set of labeled  molecular graphs $\{G_L = G_1, G_2, ...G_l\}$ with corresponding property $\{y_1,y_2,...y_l\}$ and a set of unlabeled molecular graphs $\{G_{l+1}, ..., G_{l+u}\}$, our goal is to learn a model that predicts the set of labels $\{y_{l+1},..., y_{l+u}\}$ for the unlabeled molecular graphs as part of its latent representation. This model should be able to reconstruct the molecular graph $\{G\}$ from the learned latent representation. Therefore, changing the labels will lead to generating new molecular graphs conditioned on the target labels.

\subsection{SeMole}

Our method extends Junction Tree Variational Autoencoder to a semi-supervised generative model for molecular graph property prediction. We propose a semi-supervised generative model consisting of three latent variables $z_G$, $z_T$, and $y$. Figure \ref{fig1} shows the graphical model. $y$ represents the molecular graph's high-level property, such as partially observed solubility. Following Junction Tree Variational Autoencoder, the molecular graph is decomposed to a junction tree by replacing each cluster with a node, and $z_T$ represents the tree structure and the clusters in the tree. $z_G$ represents these clusters' connectivity and how they are connected in the original graph. $z_T$ and $z_G$ are unobserved. $y$, $z_T$ and $z_G$ are sampled from a normal distribution.
\begin{equation}
\begin{array}{l}
p_{\theta}(y) = N(y|0,1) \\ \\
p_{\theta}(z_T) = N(z_T|0,1), \quad
p_{\theta}(z_G) = N(z_G|0,1) \\ \\
p_{\theta}(G | z_G, T, y) = f(G ; z_G, T, y, \theta) \\ \\ p_{\theta}(T | z_T, y) = g(T ; z_T, y, \theta)
\end{array}
\end{equation}
Although molecular graph generation - $p_{\theta}(G | z_G, T, y)$ depends on junction tree $T$, which in turn depends on $y$ already, a deliberate choice was made to make both molecular graph and junction tree generation process dependent on $y$. This is justified because some stereoisomers (and constitutional isomers) pairs can have drastically different molecular properties. It is the responsibility of the graph decoder to select the proper connectivity of the clusters that give rise to the isomer with the desired property. For that reason, latent variable y is also provided to the graph decoder.
\subsubsection{Objective}
Our goal is to approximately maximize the log-likelihood for both the tree structure and the molecular graph by maximizing a variational lower bound (ELBO) for observed and unobserved y. First, for the case where y is observed the objective function is denoted as follows:
\begin{align}
    \begin{split}
        \log p_{\theta}(G, T, y) & \geq E_{q_{\phi}(z_T, z_G| T, G, y)}\big[ \log p_{\theta}(G|T, z_G, y) \\ & \quad+ \log p_{\theta}(T|z_T, y) + \log p_{\theta}(z_T) \\ & \quad+ \log p_{\theta}(z_G) + 2 * \log p_{\theta}(y)  \\ & \quad - \log q_{\phi}(z_T|T) - \log q_{\phi}(z_G|G)\big] \\ &= -\mathcal{L}(T, G, y)
    \end{split}
\end{align}
For the case where the label corresponding to a data point is unobserved, we approximate $p_{\theta}(z_G, z_T, y| T, G)$ using a posterior function $q_{\phi}(z_G, z_T, y| T, G)$ modelled by neural networks. The detailed derivation could be found in the appendix. The unlabelled dataset objective is as follows:
\begin{align}
\begin{split}
    \log p_{\theta}(G, T) & \geq E_{q_{\phi}(z_T, z_G, y| T, G)}\big[ \log p_{\theta}(G|T, z_G, y) \\ & \quad+ \log p_{\theta}(T|z_T, y) + \log p_{\theta}(z_T) \\ & \quad+ \log p_{\theta}(z_G) + 2 * \log p_{\theta}(y) \\ & \quad - \log q_{\phi}(z_T|T) - \log q_{\phi}(z_G|G) \\ & \quad- \log q_{\phi}(y|T,G)\big] \\ & =
   \sum_{y} \big[-\mathcal{L}(T, G, y)\big] + \mathcal{H}\big(q(y|T,G)\big) \\ & = - \mathcal{U}(T, G)
    \end{split}
\end{align}

Following \cite{kingma2014semi}, it is desirable to add a loss so the distribution $q_{\phi}(y|T,G)$ can be learnt with the labelled dataset. This yields our final objective function:
\begin{align}
\label{obj} 
\begin{split}
    \mathcal{J}^a & = \sum_{(T,G,y) \sim \mathcal{D}} \mathcal{L}(T, G, y) + \sum_{(T,G) \sim \mathcal{D}} \mathcal{U}(T, G) \\ & \quad+ a \cdot E \big[\log q_{\phi}(y|T,G)\big]
\end{split}
\end{align}

The reconstruction loss follows the JTVAE\cite{jin2018junction} loss function consisting of a topological and label prediction for the tree structure and the prediction of correct subgraphs for the graph structure. However, unlike JTVAE\cite{jin2018junction}, the graph decoder and the tree decoder generate data conditioned on the target property. The last term is a mean squared error loss that is added for supervised property prediction. 

\begin{table*}
\centering
\caption{Mean Absolute Error(MAE) of Molecule Property Prediction with Varying Percentage of Labeled data}
\begin{tabular}{lllllll}
\hline \% Labeled&target property&SSVAE&SeMole$_{Pretrained}$&SeMole&SeMole$_{Supervised}$\\
\hline 5\%&MolWt&$\textbf{1.639} \pm \textbf{0.577}$&$1.658 \pm 0.130 $&$1.642 \pm 0.165$&$1.894 \pm 0.131$\\
&LogP&$0.120 \pm 0.006$&$\textbf{0.117} \pm \textbf{0.001} $&$0.133 \pm 0.002$&$0.154 \pm 0.004$\\
&QED&$0.028 \pm 0.001$&$\textbf{0.021} \pm \textbf{0.000}$&$0.038 \pm 0.000$&$0.057 \pm 0.000$\\
\hline 10\%&MolWt&$1.444 \pm 0.618$&$\textbf{1.350} \pm \textbf{0.139}$&$1.419 \pm 0.149$&$1.597 \pm 0.126$\\
&LogP&$0.090 \pm 0.004$&$0.092 \pm 0.001$&$\textbf{0.089} \pm \textbf{0.000}$&$0.127 \pm 0.006$\\
&QED&$0.021 \pm 0.001$&$\textbf{0.017} \pm \textbf{0.000}$&$0.024 \pm 0.000$&$0.046 \pm 0.000$\\
\hline 20\%&MolWt&$\textbf{1.008} \pm \textbf{0.370}$&$1.069 \pm 0.088$&$1.187 \pm 0.119$&$1.286 \pm 0.121$\\
&LogP&$0.071 \pm 0.007$&$\textbf{0.070} \pm \textbf{0.000}$&$0.076 \pm 0.000$&$0.091 \pm 0.001$\\
&QED&$0.016 \pm 0.001$&$0.012 \pm 0.000 $&$\textbf{0.011} \pm \textbf{0.000}$&$0.019 \pm 0.000$\\
\hline 50\%&MolWt&$1.050 \pm 0.164$&$\textbf{1.049} \pm \textbf{0.041}$&$1.054 \pm 0.387$&$1.243 \pm 0.273$\\
&LogP&$0.047 \pm 0.003$&$\textbf{0.043} \pm \textbf{0.000}$&$0.047 \pm 0.000$&$0.061 \pm 0.000$\\
&QED&$0.010 \pm 0.001$&$\textbf{0.009} \pm \textbf{0.000}$&$0.011 \pm 0.000$&$0.013 \pm 0.000$\\
\hline
\label{tab1}
\end{tabular}
\end{table*}

\subsection{Pretraining}
Semi-supervised Variatiatioanl Autoencoders \cite{kingma2014semi} are challenging to train end-to-end due to their multiple stochastic latent variables \cite{maaloe2016auxiliary}. However, Kingma et al. stacked a pretrained feature extractor to their method, which improved the performance significantly. Here we propose a pretraining version of SeMole called SeMole$_{Pretrained}$ by setting the coefficient of the supervised loss, $\alpha$, to zero for the first ten training epochs. Since the supervised loss decreases during the training instead of assigning $\alpha$ to a constant number, we gradually increase $\alpha$ during the training until it reaches $\alpha_{max}$, which is a hyperparamater. 

\section{Experiments}
We evaluate the effectiveness of our proposed methods for two tasks: molecular property prediction and conditionally generating molecules with desired properties. 

We vary the percentage of the labeled data (5\%, 10\%, 20\%, and 50\%) in the molecular property prediction experiments to evaluate the semi-supervised component of our method. We save 5\% of the training set for validation. The property prediction task is evaluated by Mean Absolute Error (MAE) on the testing dataset, which consists of 10000 molecules. 

Considering the prior distribution of target properties in the model, we normalize the properties to have a mean of 0 and a standard deviation of 1. We set the batch size to 16 and the learning rate to 0.001. The dimension of $z_T$ and $z_G$ is set to 56. The tree encoder consists of two GRU networks, and the graph decoder is a Message Passing Neural Network. 

We use SSVAE \cite{kang2018conditional} as the baseline which also is a semi-supervised variational autoencoder but representing molecules as SMILES representation instead of molecular graphs. The results are copied from the original paper since we use the same experimental design as our baseline. We further perform an ablation study for three different versions of our proposed model to assess the impact of semi-supervision and pretraining. 

We developed a supervised version of SeMole by eliminating the unlabeled data from training to show the impact of semi-supervised learning as a solution to label scarcity. Therefore, SeMole$_{Supervised}$ only takes the labeled portion of the training data as the input. Our goal is to show the benefits of leveraging unlabeled data compared to supervised training.

We use the trained models on 50\% labeled data to generate molecules conditioned on target properties and set the properties to different specific values. We also generate molecules unconditionally. Following our baseline \cite{kang2018conditional}, during the generation of molecules, we check the validity of the generated molecules using RDKit package \cite{Landrum2016RDKit2016_09_4}, and we discard invalid molecules, or molecules that already exist in the training data or are already generated by the decoders before. We continue this process until we generate 3000 molecules or we reach the limit of 10000 generated molecules. Then we label the generated molecules for the target properties to assess whether their properties are close to the target properties.

\subsection{Dataset}

We use 310000 drug-like molecules sampled from ZINC \cite{sterling2015zinc}. We use the same dataset that was also used to evaluate SSVAE \cite{kang2018conditional}. 
Following the literature, we use three chemical properties that are available using RDKit Package\cite{Landrum2016RDKit2016_09_4}, i.e., molecular weight (MolWt), Wildman-Crippen partition coefficient (LogP), and quantitative estimation of drug-likeness (QED) which are generaly used for this task in the literature.

\subsection{Results}

Table\ref{tab1} demonstrates the MAE of the molecule property prediction task with the varying number of labels in the training dataset. We repeated the experiment three times and reported the average and standard deviation of the MAE. The pretrained version of SeMole outperformed SSVAE in most of the cases.
SeMole$_{Pretrained}$ achieved better performance on LogP and QED compared to MolWt. The results show the effectiveness of the pretraining since SeMole$_{Pretrained}$  outperforms SeMole in all cases except in two experiments where SeMole performs slightly better. SeMole$_{Pretrained}$ substantially outperforms SeMole$_{Supervised}$ showing the benefit of semi-supervision while having limited labeled data. The performance of SeMole$_{Pretrained}$ compared to SSVAE shows the benefit of representing molecules as graphs and the pretraining of semi-supervised generative models for molecular property prediction with partial supervision. 

We compared the percentage of valid, unique, and novel molecules generated by these methods. The results show that SSVAE and SeMole$_{Pretrained}$ demonstrate better performance than SeMole$_{Supervised}$ and SeMole. The main distinction between SeMole$_{Pretrained}$ and SSVAE is that SeMole$_{Pretrained}$ generates 100\% valid molecules. SeMole$_{Pretrained}$ also outperforms SSVAE on the percentage of the generated molecules with properties within the threshold of 5\% difference from the target values. The table of generative task results is not included due to the page limit.

\section{Conclusion}

We have proposed SeMole, a semi-supervised generative model for molecular graphs. we augmented a state-of-the-art generative model, JTVAE, for molecular graphs with
semi-supervised learning. We also added pretraining phase to improve the training of the semi-supervised generative model. We performed experiments on molecular property prediction and conditional generation only using limited labeled data. These experiments were performed on three properties in the ZINC dataset. The SeMole$_{Pretrained}$ outperforms SSVAE on most molecule property prediction tasks and generates 100\% valid molecules conditioned on target properties. The ablation study showed the effectiveness of semi-supervision over supervised methods when labeled data is limited. We also demonstrated the efficacy of our pretraining phase for training a Semi-supervised VAE.

This paper suggests several directions for future research. SeMole$_{Pretrained}$, like other methods in the literature, predicts only on molecular property and generates molecules conditioned on a single property. However, predicting multiple molecular properties could improve the accuracy of the prediction and also is beneficial to discover molecules with multiple target properties. Moreover, extending the 2D representation of molecules to 3D representation could help learn richer latent representations, which lead to more accurate predictions and improved molecule generation.

\bibliography{aaai23}

\end{document}